%% file: main.tex
\documentclass[runningheads]{llncs}

\usepackage{eccv}

\usepackage{eccvabbrv}

\usepackage{graphicx}
\usepackage{booktabs}
\usepackage{amsmath}
\usepackage{amssymb}
\usepackage{enumitem}
\usepackage{amsfonts}
\usepackage{algorithm}
\usepackage{algpseudocode}
\usepackage{pifont}

\usepackage[numbers]{natbib}
\usepackage{dblfloatfix}
\usepackage{longtable}
\usepackage{listings}
\usepackage{xcolor}
\usepackage{longtable}

\definecolor{codegreen}{rgb}{0,0.6,0}
\definecolor{codegray}{rgb}{0.5,0.5,0.5}
\definecolor{codepurple}{rgb}{0.58,0,0.82}
\definecolor{backcolour}{rgb}{0.95,0.95,0.92}

\lstdefinestyle{mystyle}{
    backgroundcolor=\color{backcolour},   
    commentstyle=\color{codegreen},
    keywordstyle=\color{magenta},
    numberstyle=\tiny\color{codegray},
    stringstyle=\color{codepurple},
    basicstyle=\ttfamily\footnotesize,
    breakatwhitespace=false,         
    breaklines=true,                 
    captionpos=b,                    
    keepspaces=true,                 
    numbers=left,                    
    numbersep=5pt,                  
    showspaces=false,                
    showstringspaces=false,
    showtabs=false,                  
    tabsize=2
}
\input{macro.tex}

\usepackage[accsupp]{axessibility}  

\usepackage{hyperref}

\usepackage[capitalize]{cleveref}

\usepackage{orcidlink}

\begin{document}

\title{HARIVO: Harnessing Text-to-Image Models for Video Generation} 

\titlerunning{HARIVO}

\author{Mingi Kwon\inst{1,2,5}, Seoung Wug Oh\inst{2}, Yang Zhou\inst{2}, Difan Liu\inst{2}, Joon-Young Lee\inst{2}, Haoran Cai\inst{2}, Baqiao Liu\inst{2,3}, Feng Liu\inst{2,4}, Youngjung Uh\inst{1}}

\authorrunning{Kwon et al.}

\institute{Yonsei University \and
Adobe\footnote{This work was done during the internship at Adobe} \and
University of Illinois Urbana-Champaign \and
Portland State University \and
GivernyAI
}


    \maketitle
    \begin{center}
        \includegraphics[width=\textwidth]{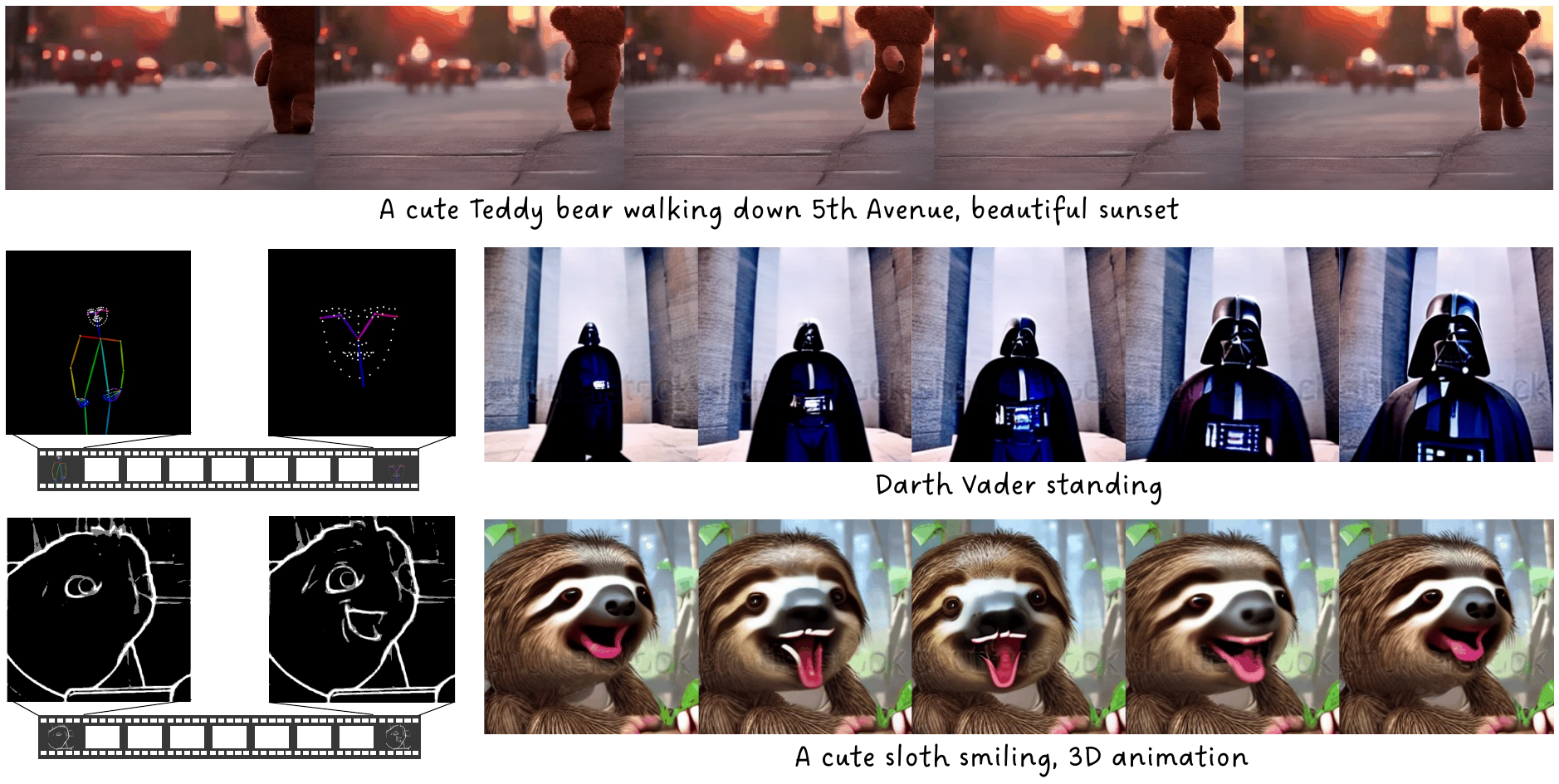}
        \captionof{figure}{Our method generates high quality videos for given text prompts and is easy to combine with other methods such as ControlNet.}
        \label{fig:teaser}
        \vspace{-1em}
    \end{center}

\begin{abstract}
We present a method to create diffusion-based video models from pretrained Text-to-Image (T2I) models. Recently, AnimateDiff proposed freezing the T2I model while only training temporal layers. We advance this method by proposing a unique architecture, incorporating a mapping network and frame-wise tokens, tailored for video generation while maintaining the diversity and creativity of the original T2I model. Key innovations include novel loss functions for temporal smoothness and a mitigating gradient sampling technique, ensuring realistic and temporally consistent video generation despite limited public video data. We have successfully integrated video-specific inductive biases into the architecture and loss functions. Our method, built on the frozen StableDiffusion model, simplifies training processes and allows for seamless integration with off-the-shelf models like ControlNet and DreamBooth. project page: \href{https://kwonminki.github.io/HARIVO/}{https://kwonminki.github.io/HARIVO/}
\end{abstract}

\section{Introduction}
\label{sec:intro}
Recently, diffusion-based Text-to-Image (T2I) models have shown unprecedented performance \cite{rombach2022high,nichol2021glide,ho2022imagen,balaji2022ediffi}. Upon this success, there are many attempts in the community to design diffusion-based Text-to-Video (T2V) models. A common approach is to introduce additional 3D convolution and temporal attention blocks to the T2I model and fine-tune the whole network on videos in the same way as for T2I models on images \cite{ho2022video,molad2023dreamix,zhou2022magicvideo,he2023latent,ge2023preserve,gafni2022make,wang2023modelscope}. However, these whole-training T2V models have several disadvantages. First, training the entire model requires a large amount of data. Therefore, most video models have used additional in-house datasets. Second, the models diminish the variety of styles due to the highly realistic nature of training videos. To prevent this, a joint-training approach with images and videos is being used, however, we believe this does not tackle the root of the problem effectively. Additionally, most models consist of multiple stages, e.g., keyframe generation and spatial/temporal interpolation, and have the disadvantage of struggling to maintain temporal consistency due to sparse keyframe generation.


Training only the temporal components while keeping the T2I model parameters frozen is an alternative that avoids the aforementioned problems and reduces training complexity~\cite{blattmann2023align, guo2023animatediff}. By freezing the T2I model, we can anticipate diverse outcomes based on the knowledge trained from images, and using existing methods from image models. \textit{However, unfortunately, training only the temporal components is not straightforward and often does not converge easily.} In response, previous studies have adopted the following methods.

VideoLDM\cite{blattmann2023align} continued to adopt multi-stage training with spatial/temporal interpolation and upsampling, which could not reduce the training complexity. Furthermore, instead of directly using the image model, adapting the masked blend T2I feature strategy makes it difficult to combine with existing T2I methods. AnimateDiff \cite{guo2023animatediff} mainly generates personalized videos assisted by Dreambooth~\cite{ruiz2023dreambooth} and LoRA~\cite{hu2021lora}, which facilitate the creation of videos by generating similar images; but struggles in generating diverse videos.

\begin{figure*}[!t]
    \centering
    \includegraphics[width=1\linewidth]{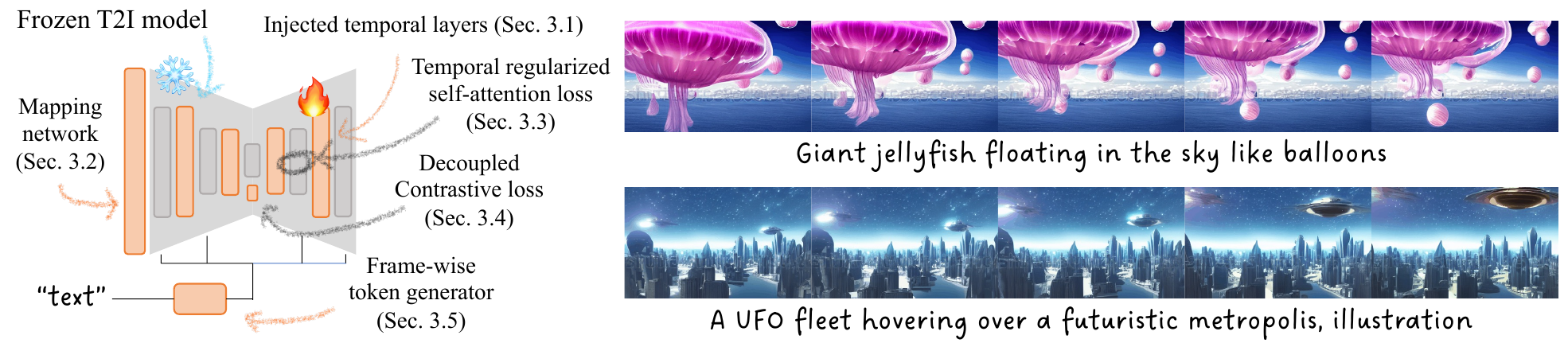}
    \vspace{-2em}
    \caption{An overview of our method: illustration and results. We propose a method where we freeze a Text-to-Image model and train only temporal layers, including the additional proposed networks. This technique enables the successful generation of videos in various styles.
    }
    \vspace{-2em}
    \label{fig:concept}
\end{figure*}

In this paper, we demonstrate that when freezing the image model and only training the temporal layers, we can employ additional inductive biases that we know about videos. We propose a collection of novel designs, including a novel architecture, training losses, and a sampling technique that enable us to fully leverage the power of a frozen T2I model. It generates videos of high quality and diverse styles, similar to the image model, and maintains temporal consistency as shown in \fref{fig:teaser} and \ref{fig:concept}.

Our approach offers several advantages. Firstly, it supports the combination of personalized T2I models such as DreamBooth, LoRA, ControlNet\cite{zhang2023adding}, and IP-Adapter\cite{ye2023ip-adapter} without requiring further training to produce personalized videos. This goes beyond merely using off-the-shelf models to show that one can design methods and train models using only images to control videos. Secondly, our model shows temporally consistent video generation although it is trained on a \textit{public dataset} (WebVid-10M), whereas many existing works are trained on in-house datasets. Thirdly, we introduce various independent methods that apply the many features of image models to the training of video models. We extend the ability of T2I models to the video domain with minimal modification. All of our components bring orthogonal improvements, enabling other researchers to selectively use them. 

\paragraph{Extending T2I model to Video model.} We propose two additional components that help to extend the T2I model to a video model: a \textit{mapping network} and \textit{frame-wise tokens}. Since T2I models are originally trained to generate a variety of independent images from randomly sampled Gaussian noises, we conjecture that generating videos from independent and identically distributed (IID) noises for all frames is not suitable because consecutive frames of videos are highly correlated.
To address this, we introduce a \textit{mapping network}, an additional block positioned before the UNet and implicitly adjusting the noise distribution to facilitate video generation. This technique is motivated by the discovery that the low frequency component is largely determined by initial noise \cite{khrulkov2022understanding,meng2021sdedit,choi2021ilvr,wu2022unifying}. In addition, we introduce \textit{frame-wise tokens} which handle similiar but subtle differences between frames. For this, we design a token generator that receives the text prompt and produces frame-wise tokens. The frame-wise tokens are concatenated to the original text tokens such that our T2V model generates frames with the same content specified by the text prompt and slightly different dynamic states.

\paragraph{Losses for temporal smoothness.} 
In addition to the model improvements, we propose two loss functions that help to generate temporally consistent videos. 
First, \textit{temporal regularized self-attention loss} penalizes the difference in the self-attention map between frames. Inspired by early studies that show
the crucial role of self-attention maps in specifying scene structure and object identity \cite{tumanyan2022plug,cao2023masactrl,khachatryan2023text2video}, it encourages smoothness between frames, improving temporal consistency and coherency.
Second, \textit{decoupled contrastive loss on h-space} ensures all frames share the same semantic content. As the bottleneck feature (h-space) of U-Net has semantics \cite{kwon2022diffusion,park2023understanding,jeong2023training}, we impose the h-space features of all frames within a video to be close to each other and far from different videos.
\paragraph{Mitigating gradient sampling.}
Lastly, but importantly, we propose \textit{mitigating gradient sampling} that prevents abrupt changes between frames for each timestep at inference time. It provides guidance during the generative process from the gradient of the difference between frames. It helps to generate realistic videos that are naturally continuous.

\paragraph{Experiments and contribution.}
We build our video model upon the frozen StableDiffusion, a widely-used public T2I model, and train the mapping network, frame-wise token generator, and temporal layers with WebVid-10M dataset. We provide an ablation study and comparison with Fréchet Video Distance (FVD)\cite{unterthiner2019accurate} on UCF101 \cite{soomro2012ucf101} and Clip Similarity \cite{radford2021learning} on MSR-VTT \cite{xu2016msr-vtt} following previous works. 
Our contributions are 1) novel and well-motivated additional components for video model, 2) single stage training with public dataset, 3)~showing temporal consistency, 
and 4) easy combination with off-the-shelf personalized T2I models (DreamBooth, LoRA, ControlNet, and IP-Adapter).


\section{Background and Related Work}
Diffusion models (DMs) have demonstrated impressive performance in the realm of image synthesis. The fundamental mechanism of DMs involves a predetermined forward diffusion process, where noise is incrementally introduced to the original data $\vx_0 \sim p(\boldsymbol{x}_0)$, alongside the training of a denoising model to invert this procedure through a denoising score matching loss expressed as:
\begin{equation}
L_{\text{simple}} = \mathbb{E}_{t, \vx_0, \boldsymbol{\epsilon}}\left|\boldsymbol{\epsilon}-\mathbf{s}_\theta\left(\vx_t, t\right)\right|_2^2 .
\label{eq:simple}
\end{equation}
DMs define the forward process as $d \mathbf{x}=\mathbf{f}(\mathbf{x}, t) d t+g(t) d \mathbf{w}$, where $\mathbf{w}$ is the Wiener process. The corresponding reverse process is: $d \mathbf{x}=[-\mathbf{f}(\mathbf{x}, T-t)+ \left.g(T-t)^2 \nabla_{\mathbf{x}} \log p_{T-t}(\mathbf{x})\right] d t+g(T-t) d \mathbf{w}$ where DMs approximate $\nabla_{\mathbf{x}} \log p_{t}(\mathbf{x})$ with $s_{\theta}(\mathbf{x},t)$.

In light of the intricate challenges posed by high-dimensional data such as RGB images, the Latent Diffusion Model (LDM) \cite{rombach2022high} has been introduced. LDM simplifies the task by conducting the diffusion and denoising procedures within a low-dimensional latent space derived from a Variational Autoencoder (VAE). For the sake of brevity, we choose to exclude the specific notations related to the Encoder and Decoder in this paper.

Following the advancement of T2I, the research community has pioneered Text-to-Video (T2V). Most T2V models have evolved from the 3D UNet structure proposed by VideoDiffusion\cite{ho2022imagen}, training with multi-stage 3D UNets through complete finetuning. Imagen Video\cite{ho2022video} and Make-A-Video\cite{singer2022makeavideo} have successfully demonstrated multi-stage T2V models at the RGB pixel level, while few works\cite{zhou2022magicvideo, wang2023modelscope} integrated temporal layers into the multi-stage T2I LDM.

VideoLDM\cite{blattmann2023align} first presented a method to train T2V models while keeping the T2I models frozen, but applying a masked blending method (not using the T2I model's features directly), and employing multi-stage training techniques. AnimateDiff\cite{guo2023animatediff} advanced this by keeping the T2I entirely frozen and only training the temporal layer in one stage, but AnimateDiff did not generalize successfully to a text-to-video model, being limited to specific video domains using Dreambooth or LoRa.

In contrast, our approach, unlike other studies that solely relied on basic denoising score matching loss, proposes new losses and architectures. We have successfully trained the video model while keeping the T2I model completely frozen, enabling flexible combinations with multiple existing off-the-shelf models.

\section{Method}
We briefly describe inflating text-to-image (T2I) model to video model as a baseline and then introduce our novel components. Please note that in all the figures within the paper, the parts highlighted in orange indicate the network components that we are training.

\subsection{Baseline video model}
Similar to AnimateDiff \cite{guo2023animatediff}, we add temporal attention layers \temporal{} between spatial layers \spatial{} of the text-to-image model, as a common technique for video generation methods.

\subsection{Mapping network}
\label{sec:method_mapping}
In diffusion models, training and sampling rely on Gaussian noises $\boldsymbol\epsilon$ and $\vx_T \in \mathbb{R}^{B \cdot F \times C \times H \times W}$,
where $B, F, C, H$ and $W$ are batch size, number of frames, number of channels, height, and width, respectively. Some previous methods with pretrained T2I models design specific distributions, different from Gaussian, to model the relationship between frames in a video \cite{ge2023preserve,blattmann2023align}. For example, PYoCo used a correlated noise model that employs specific shared noise across all frames to design frame correlations. While this approach fits a Gaussian distribution on a frame-wise basis, it does not conform to a Gaussian distribution across the entire video space of $\mathbb{R}^{B \cdot F \times C \times H \times W}$ dimensions (See Appendix). Disregarding the Gaussian distribution nullifies the reparametrization trick of diffusion models, leading to unstable training; experimentally, we have found that models always diverge when training only on video without using joint training with images\footnote{
When employing joint training, each frame satisfies the properties of being independent and identically distributed (IID), which prevents the aforementioned issue.}.


Therefore, we introduce a \textit{mapping network} \mapping{} that transforms the diffusion noise prior (IID noise) into a distribution more suitable for generating videos.
Unlike previous methods, it learns the mapping from Gaussian distribution to an implicit distribution for video supervised by the denoising score matching loss (\eref{eq:simple}). In addition, it ensures the validity of the reparametrization trick for training. 
We design the mapping network to have temporal attention and 3D convolutions such that the inter-frame interaction makes it easier to prepare the input for the Video U-Net to match the relationship between frames within a video. Please refer to Appendix for detail.

\begin{figure}[!t]
    \centering
    \includegraphics[width=1\linewidth]{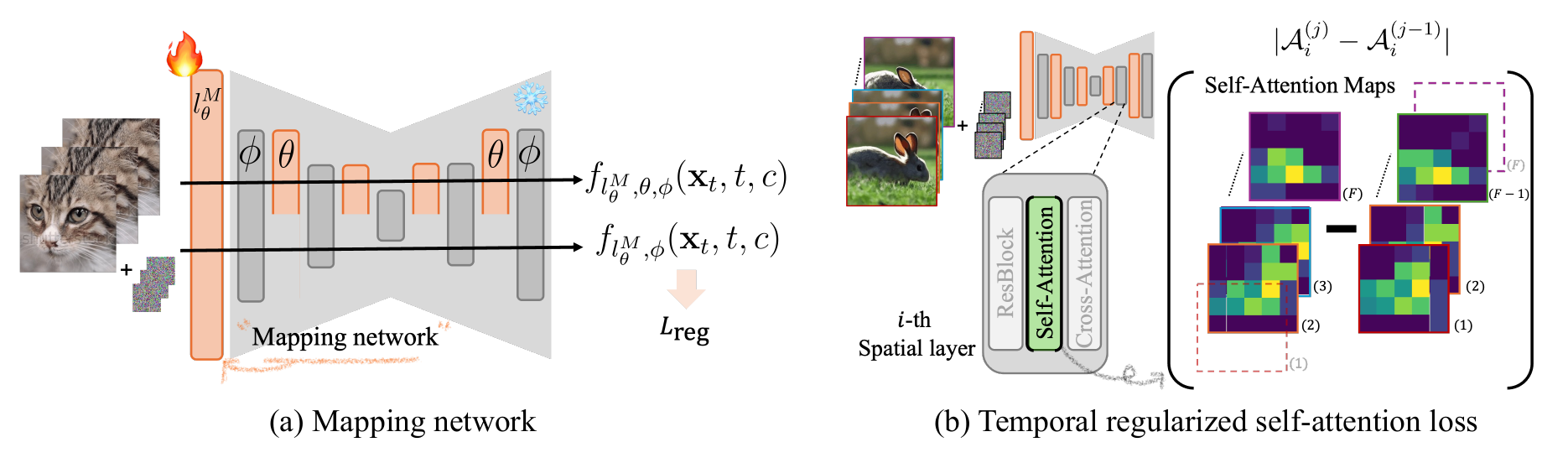}
    \vspace{-2em}
    \caption{
     \textbf{(a) Our mapping network} in front of the U-Net maps IID Gaussian noise to a proper distribution for video generation. \textbf{$L_\text{reg}$} penalizes difference between frames with and without temporal layers $l_\theta$ to preserve the expertise of the pretrained T2I model. \textbf{(b) Temporal regularized self-attention loss} penalizes the difference between self-attention maps of adjacent frames in the frozen T2I model to improve smoothness between consecutive frames.
    }
    \vspace{-2em}
    \label{fig:mapping}
\end{figure}

To prevent the mapping network from pushing the distribution to excessively far from the original, we introduce a \textit{regularization} to safeguard the T2I model's expertise:
\begin{equation}
    L_{\text{reg}} = \lambda_t ||f_{l^{M}_{\theta}, \theta, \phi}(\mathbf{x}_t,t,c) - f_{l^{M}_{\theta}, \phi}(\mathbf{x}_t,t,c)||^2_2 ,
    \label{eq:reg}
\end{equation}
where $f_{l^{M}_{\theta}, \theta, \phi}$ denotes the whole T2V model,$f_{l^{M}_{\theta}, \phi}$ denotes T2I model with the mapping network, and $c$ denotes the input text prompt. The T2I model individually denoises different frames.
\fref{fig:mapping}(a) illustrates the mapping network and above regularization.
It guides the output of the mapping network to lie in the domain of T2I model so that it becomes easier for the temporal layers in T2V model to work with the frozen T2I model.
Furthermore, it also encourages the T2V model to preserve the existing knowledge of the T2I model similariy to distillation.
We set $\lambda_t = 1 - \frac{t}{T}$ to soften the regularizer on timesteps close to $T$ because T2I model produces discontinuous frames in early timesteps. It ensures that, at low noise levels, the image model's output closely aligns with the ground truth, matching the video model's output.

\subsection{Temporal regularized self-attention loss}
The self-attention map within diffusion models encapsulate structural content of the generated images \cite{kwon2022diffusion,tumanyan2022plug,jeong2023training,cao2023masactrl,park2023understanding}.
Upon this understanding of self-attention maps, Text2video-zero\cite{khachatryan2023text2video} and MasaCtrl\cite{cao2023masactrl} pioneers generating videos by sharing the self-attention map without additional training.

Taking inspiration from their approaches, we propose the \textit{Temporal Regularized Self-attention} (TRS) loss, which aims to suppress the difference of self-attention maps between consecutive frames:
\begin{equation}
L_{\text{TRS}} = \sum_{i}^{N} \sum_{j=2}^{F} \lambda_{{i}} | \mathcal{A}^{(j)}_{{i}} - \mathcal{A}^{(j-1)}_{{i}} |,
\end{equation}
where $\mathcal{A}^{(j)}_{{i}}$ denotes self-attention map of $i$-th layer on $j$-th frame. We use $\lambda_{i}=\frac{i}{N}$ where 
$N$ is the total number of decoder layers to put more weights on later layers.
We compute \trs{} only on the decoder layers in U-Net.
\fref{fig:mapping}(b) depicts the TRS loss.

Note that TRS loss supervises temporal layers and the T2I layers are kept frozen.
However, although the TRS loss effectively encourages temporal coherence by preventing extreme changes, it only considers consecutive frames. In the next section, we introduce a loss function to maintain temporal consistency within a video, preventing objects from diverging to another object.




\subsection{Decoupled contrastive loss on h-space}
\label{sec:contrastive}
\begin{figure}[!t]
    \centering
    \includegraphics[width=1\linewidth]{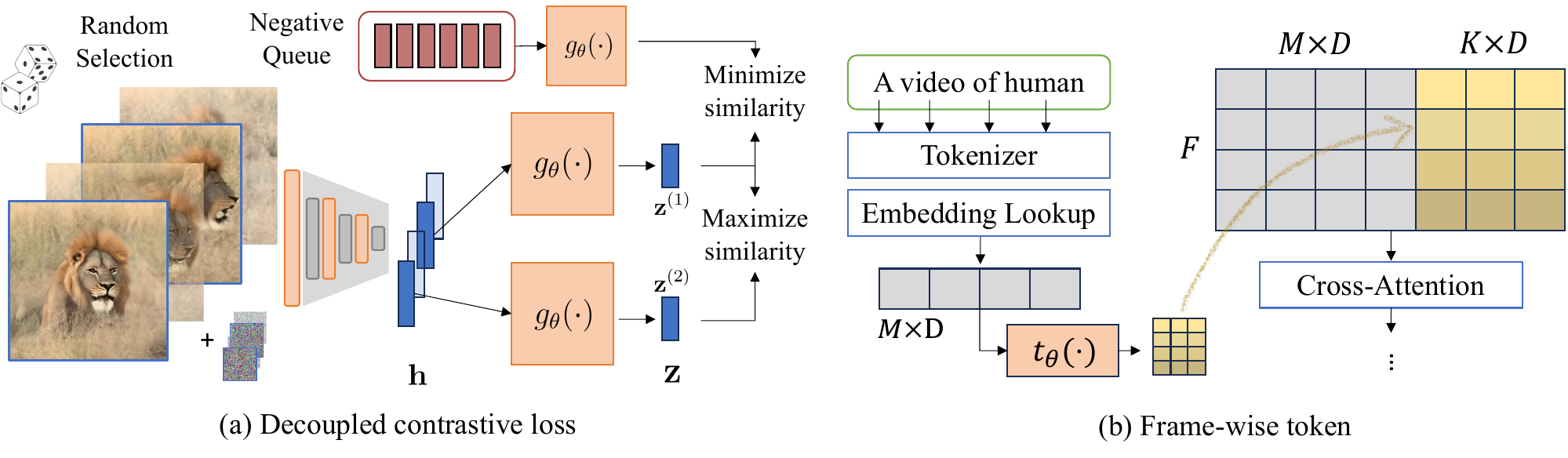}
    \vspace{-2em}
    \caption{
    \textbf{(a) Decoupled contrastive loss on \vh{}-space} encourages semantic consistency within a video. \vh{}-space is the bottleneck of U-Net. Positive pairs are randomly chosen from a video and negative samples are stored in a queue. $g_\theta$ is a projection layer used only for training.
    \textbf{(b) Frame-wise token generator} $t_{\theta}$ produces frame-wise tokens that represent subtle difference across frames. These tokens are concatenated to the text tokens to be fed into the cross-attention layers.
    }
    \vspace{-2em}
    \label{fig:contrastive}
\end{figure}

Asyrp\cite{kwon2022diffusion} shows that the deepest feature maps of UNet, named \vh{}-space, contains semantic information regardless of the sampling timestep $t$. Thus, assuming that all frames in a natural video have semantically consistent objects, we propose a loss to make all frames have a similar \vh{} using the decoupled contrastive loss \cite{yeh2022decoupled} as shown in \fref{fig:contrastive}(a).

We define positive pairs by randomly sampling two frames from a video and extracting their deepest feature maps \vh{}'s.
We store negative samples from other videos in a queue and update the queue at every step with previously-used positive pairs.
With these pairs, we employ an additional projection layer $g_{\theta}(\cdot)$ to compute $\mathbf{z} = g_{\theta}(\mathbf{h})$. Finally, we obtain the decoupled contrastive loss \cite{yeh2022decoupled}:
\begin{equation}
   L_{\text{DC}}=-\left\langle\mathbf{z}^{(1)}, \mathbf{z}^{(2)}\right\rangle / \tau + \log \sum_{q \in Q} \exp \left(\left\langle\mathbf{z}^{(1)}, \mathbf{z}^{(q)}\right\rangle / \tau\right),
\end{equation}
where $\mathbf{z}^{(1)}$ and $\mathbf{z}^{(2)}$ are projected positive pairs, $Q$ is the negative queue, and $\tau$=$0.1$ is a temperature parameter.
Intuitively, $L_{\text{DC}}$ encourages all frames in a video to be closer together in \vh{}-space while pushing dissimilar frames from other videos apart.
Positive pairs can contain non-consecutive frames because \vh{}-space does not contain structure of the scene a lot. The projection layer will be discarded after training.

\subsection{Frame-wise text embedding tokens}
Text tokens condition the generative process and several works have used text tokens for personalization \cite{hertz2022prompt,gal2022textual,mokady2023null}.
Especially, HiPer \cite{han2023highly} optimizes extra tokens after the end-of-sentence token to represent visual details of the input image that are not captured by typical text tokens. If we invert multiple frames of a video, there will be extra tokens for individual frames.

In this light, we consider frames in a video as temporal variations of an image and introduce additional frame-wise tokens to represent varying details across frames. We employ a frame-wise token generator $\boldsymbol{t_\theta}: (M \times D) \rightarrow (F \times K \times D)$ that receives the original text tokens from the input text prompt to produce frame-wise tokens, where $M$ is the maximum token length (=77), $D$ is the token embedding dimension (=768), $F$ is the number of frames, and $K$ is the number of frame-wise tokens (=3). The frame-wise tokens are concatenated to the text tokens in token-dimension and fed into the cross-attention layers. Figure \ref{fig:contrastive}(b) depicts this process.

\subsection{Training}

We train the temporal layers with all the proposed losses, including the mapping network, the projection layer for the contrastive loss, and the frame-wise token generator. The trainable and frozen components are marked in orange and gray, respectively, in the figures. The final loss becomes
\begin{equation}
    L_{\text{all}}(\theta) = L_{\text{simple}} + \lambda_{\text{TRS}} L_{\text{TRS}} + \lambda_{\text{reg}} L_{\text{reg}} + \lambda_{\text{DC}} L_{\text{DC}}.
\end{equation}
Note that we do not compute $L_{\text{simple}}$ for $f_{l^{M}_{\theta}, \phi}(\mathbf{x}_t,t,c)$. All losses can reduce the movement in the video. However, interestingly, we have found through experimentation that the appropriate use of loss actually helps in generating videos with greater movement. We speculate that appropriate regularization helps the model to find a better local minimum.

\subsection{Inference with mitigating gradient sampling}
Even with a well-trained model, we observe phenomena where the structure collapses or artifacts appear and disappear when generating complex actions or fast motions. We argue that this can be effectively prevented by adding simple guidance during the generative process.

In the pursuit of smoother image sequences, we measure the similarity between consecutive frames using a kernel function \cite{lowe1995similarity}, which is integrated into a process designed by ADM\cite{dhariwal2021diffusion}. They innovated by introducing `Classifier guidance' into the generative process, with the equation 
\begin{equation}
d \mathbf{x}=\left[-\mathbf{f}\left(\mathbf{x}, t^{\prime}\right)+g\left(t^{\prime}\right)^2\left(s_\theta\left(\mathbf{x}, t^{\prime}\right)+\alpha \nabla_{\mathbf{x}} \log \Phi_t \right)\right] d t+g\left(t^{\prime}\right) d \mathbf{w},
\end{equation}
where $t^{\prime}=T-t$ signifies the reverse timestep, $\Phi$ is an extra function for the guidance, and $\alpha$ is a parameter that scales the guidance.
This guidance can be characterized in our design of $\log \Phi_t$, where each element can be seen as the similarity of the corresponding consecutive frames as expressed in the well-established literature \cite{laio2002escaping,barducci2008well}:
\begin{equation}
    \nabla_{\mathbf{x}_i} \log \Phi_t\left(\mathbf{x}\right) \leftarrow \nabla_{\mathbf{x}_i} \omega \sum_{j=2}^{F} \exp \left(-\frac{\left\|\hat{\vx_t}^{(j)}-\hat{\vx_t}^{(j-1)}\right\|^2}{2 \sigma^2}\right),
\end{equation}
where $\hat{\vx_t}^{(i)}$ denotes predicted $\vx_0$ of frame $i$ at timestep $t$ and $\sigma^2$ denotes the variance of $\hat{\vx_t}^{(j)}-\hat{\vx_t}^{(j-1)}$.



Intuitively, $\log \Phi_t$, which we refer to as mitigating gradient (MG) sampling, iteratively reduces the differences between frames during inference. If the generated videos are already smooth, its impact is minimal, but MG sampling significantly helps in preventing momentary artifacts and excessively rapid changes. We set $\omega$ as $\sqrt{(1 - \Bar{\alpha_t})/\Bar{\alpha_t}}$. The cost of MG sampling is negligible and also it impacts greatly when near noise by $\omega$. We provide the MG sampling algorithm in Algorithm \ref{al:mg}. Specifically, the median and the logarithm are used to ensure robustness against outliers in frame differences and the smoothing effect of varying frame lengths.

\begin{algorithm}[t]
\caption{Mitigating Gradient Sampling}
\label{al:mg}
\begin{algorithmic}[1]
\State $\epsilon_{\text{pred}}=s_{\theta}(\vx_t,\varnothing) + \omega_{\text{CFG}}(s_{\theta}(\vx_t,c)-s_{\theta}(\vx_t,\varnothing))$
\State $\mathcal{D} \gets \hat{\vx}_t[1:] - \hat{\vx}_t[:-1]$ \quad\quad\quad\quad\quad\quad\quad\quad\quad\quad // $\hat{\vx}_t=({\vx}_t-\sqrt{1-\alpha_t}\epsilon_{\text{pred}})/\sqrt{\alpha_t}$
\State $\mathcal{S} \gets \text{norm}(\mathcal{D}, 2).median()^2 / \log(F - 1)$ \quad\quad\, // $F$ is number of frmaes
\State $\mathcal{G}_{\phi} \gets 2 \cdot \exp(- (\text{norm}(\mathcal{D}, 2)^2 / \mathcal{S})) \cdot \mathcal{D} / \mathcal{S} \cdot \omega$ \quad\, // $\omega=\sqrt{(1 - \Bar{\alpha_t})/\Bar{\alpha_t}}$
\State $\epsilon_{\text{pred}} \gets \epsilon_{\text{pred}} + \alpha \cdot \text{concat}(0,\mathcal{G}_{\phi})$ 

\end{algorithmic}
\end{algorithm}

\begin{figure*}[!ht]
    \centering
    \includegraphics[width=0.8\linewidth]{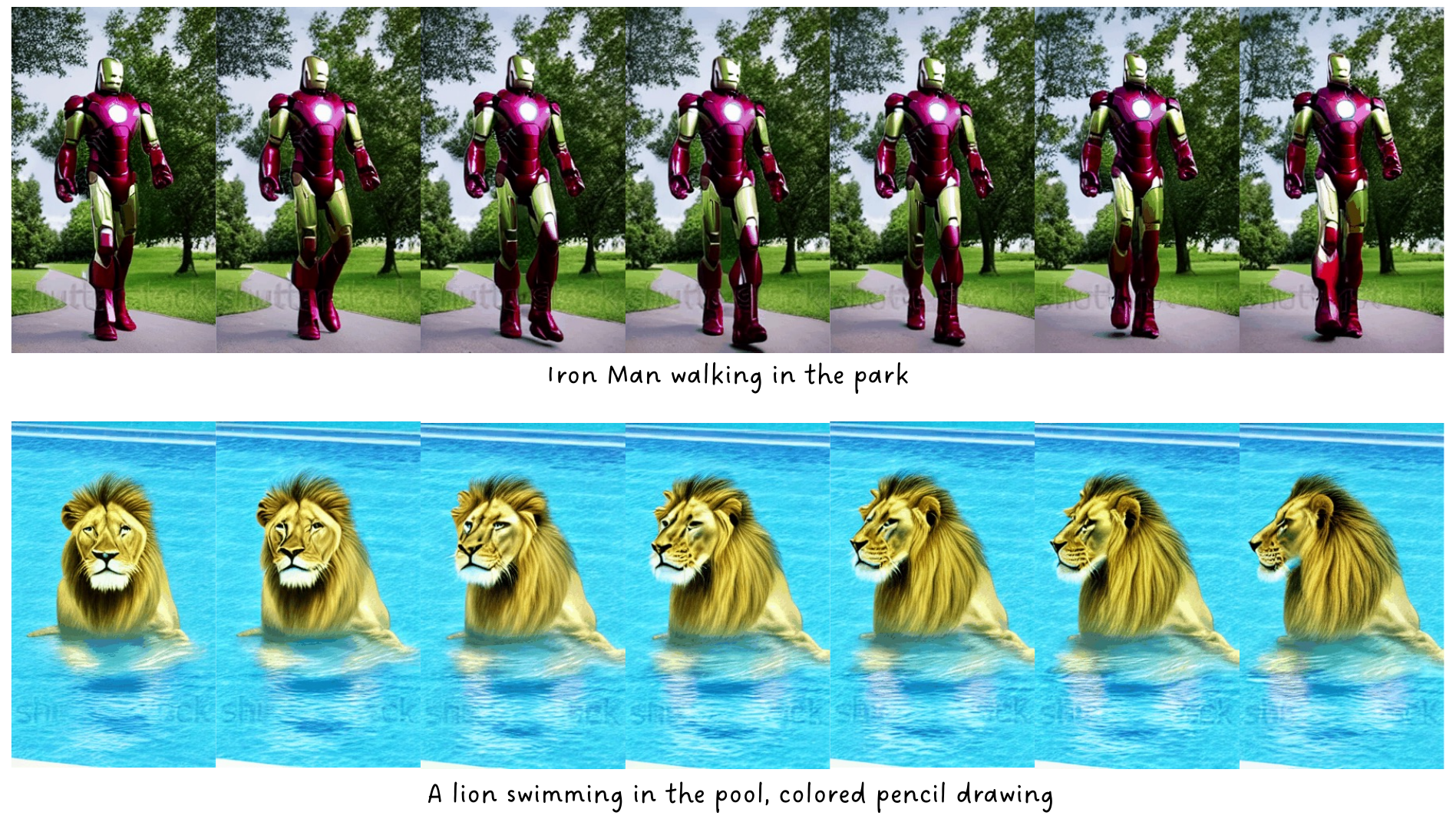}
    \centering
    \includegraphics[width=0.8\linewidth]{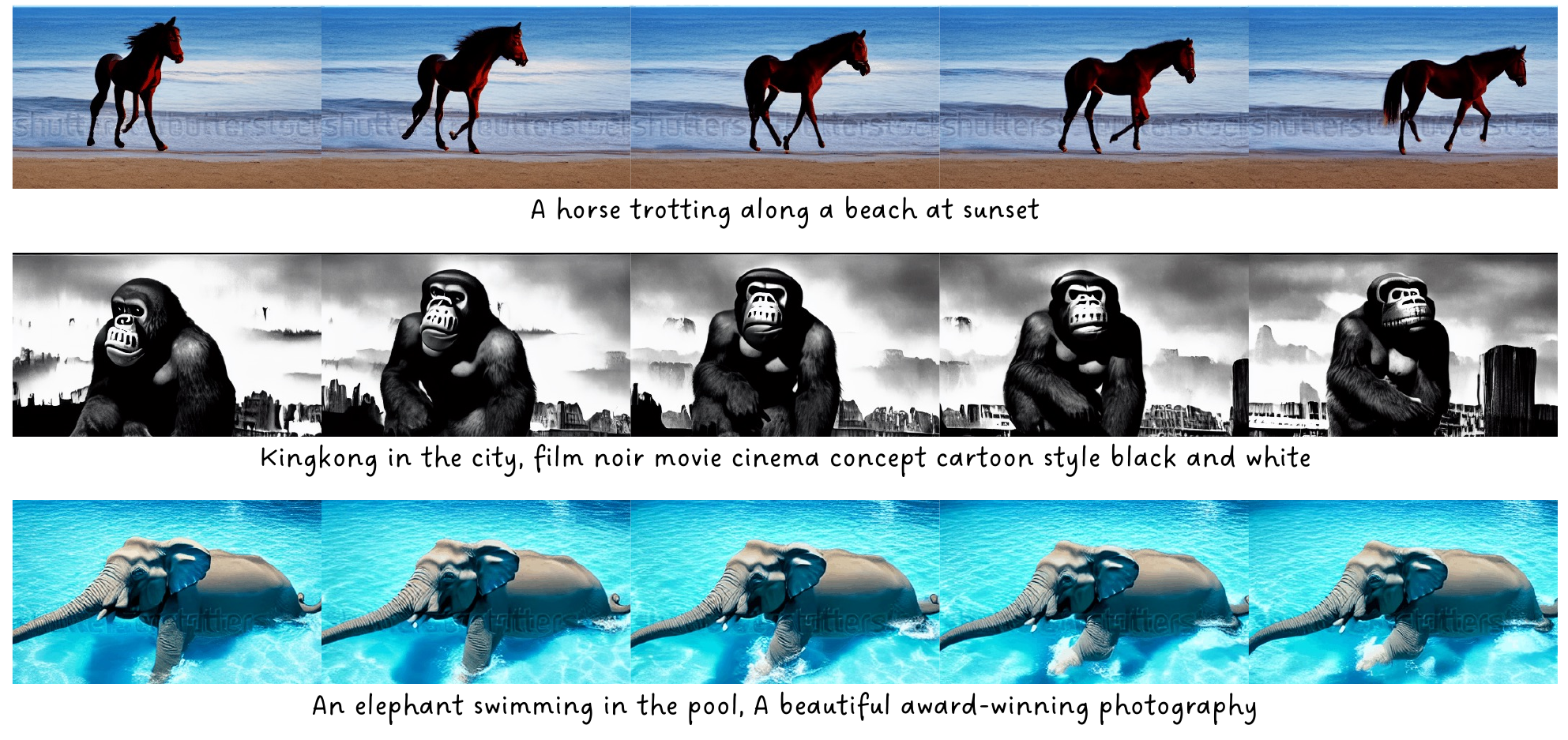}
    \centering
    \includegraphics[width=0.8\linewidth]{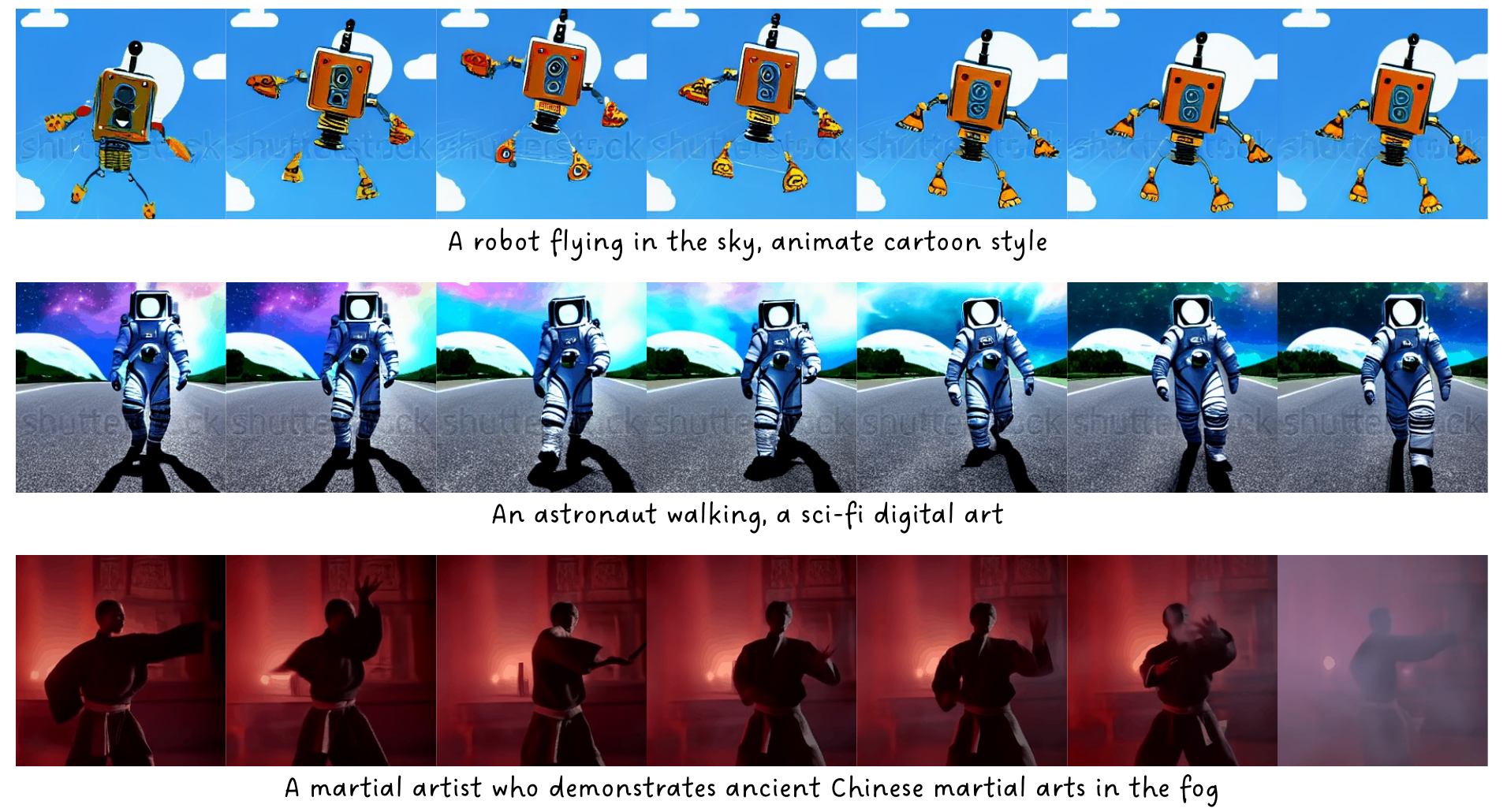}
    \caption{
    Our method generates high quality videos for given text prompts up to 512$^2$ resolution.
    }
    \label{fig:show_all}
    \vspace{-2em}
\end{figure*}

\section{Experiments}
We present thorough experiments including representative videos, ablation study, application to personalized or image-conditioned videos, and comparison to existing methods.

\paragraph{Configuration}
Our model is built upon pretrained text-to-image (T2I) StableDiffusion v1.5. After inflating it to be text-to-video (T2V) model with our method, we train the inflated part on WebVid-10M \cite{Bain21} while the T2I part is kept frozen. The videos are 4 seconds long and 6 frames per second (fps), thus 24 frames in total.
We set $\lambda_{\text{TRS}}=\lambda_{\text{reg}}=\lambda_{\text{DC}}=0.1$ and $\alpha=40$. Other details are in Appendix. For our quantitative evaluation, we use Fréchet Video Distance (FVD) \cite{unterthiner2019accurate} on UCF101 \cite{soomro2012ucf101} and Clip
Similarity \cite{radford2021learning} on MSR-VTT \cite{xu2016msr-vtt} to draw comparisons with other prevalent methods following the common practice.
On UCF101, we generate 100 videos for each class using the class names as text prompts.
On MSR-VTT, we select the first caption of each video in the test set to produce 2,990 videos.

\subsection{Qualitative results}

\fref{fig:show_all} gives example results produced by our model. Leveraging frozen StableDiffusion in its original form has allowed us to generate high-resolution videos up to $512^2$ resolution and creative videos with diverse style: watercolor painting, pencil drawing, and sci-fi digital art.  See the project page videos for details.

\subsection{Ablation study}
\tref{tab:ablation} demonstrates that our proposed components cumulatively improve Fréchet Video Distance (FVD), starting from the baseline.
We adopt AnimateDiff~\cite{guo2023animatediff} as our baseline which also trains only the temporal layers while the T2I model is frozen as our method.
Notably, the Temporal Regularized Self-attention loss ($L_\text{TRS}$) and the Decoupled Contrastive loss ($L_\text{DC}$) contribute to a significant improvement.

\begin{table}[!t]
\centering
\begin{tabular}{l|c}
                      & FVD $(\downarrow)$ \\ \hline
\quad\quad Baseline (AnimateDiff) & 1934.72  \\
{[A]} \quad + $L_{\text{TRS}}$ \& $L_{\text{DC}}$           & 1099.33   \\
{[B]} \quad + Mapping network     & 1001.41   \\
{[C]} \quad + Frame-wise tokens          & 922.67   \\
{[D]} \quad + MG Sampling (full)        & \textbf{787.87}  
\end{tabular}
\caption{\textbf{Ablation study} of the proposed components. All variants are trained for the same number of iterations from scratch.
}
\vspace{-1em}
\label{tab:ablation}
\end{table}

\begin{figure}[!t]
    \centering
    \includegraphics[width=1\linewidth]{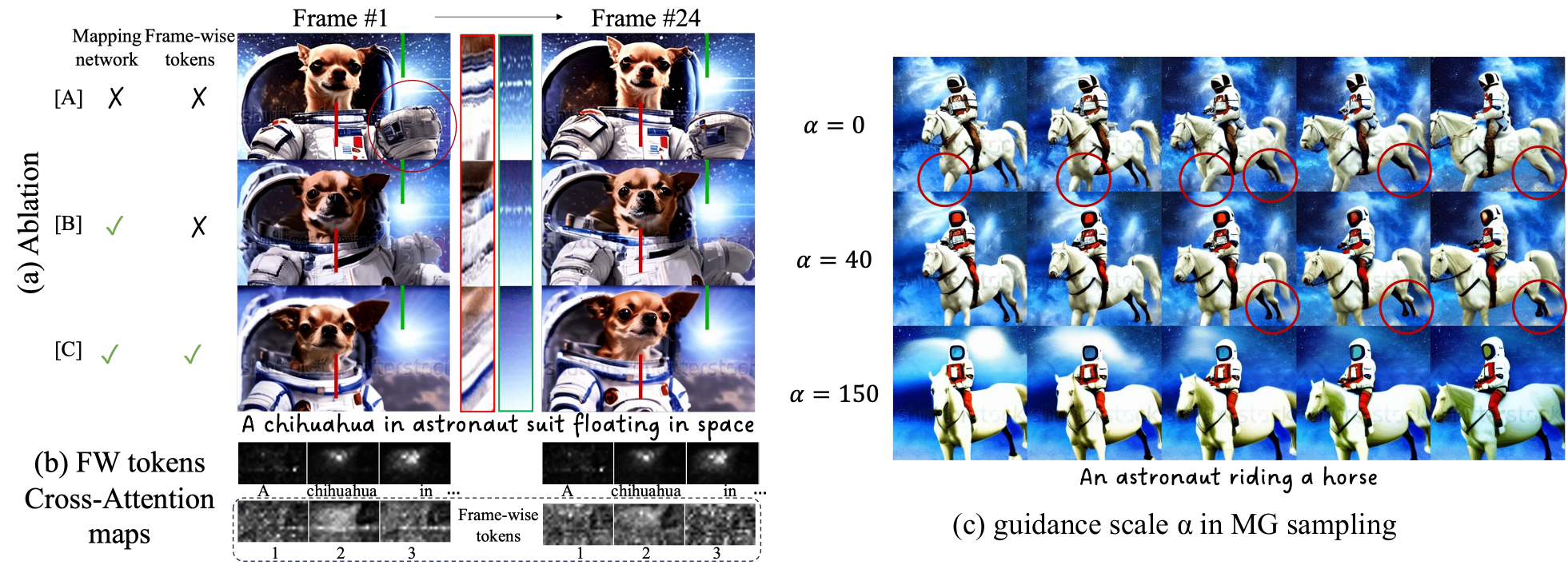}
    \vspace{-1em}
    \caption{
    \textbf{Qualitative ablation study and cross-attention maps.} The center columns are horizontal stacks of short vertical segments of fixed locations marked in red and green from 24 frames of each video.
    (a) Lacking a mapping network and frame-wise tokens causes flickering and spurious objects. 
    Using both components produce natural videos with consistent and clear objects.
    (b) Cross-attention maps for frame-wise tokens emphasize areas activated around the main object, showing frame variations. (c) Effect of varying the guidance scale $\alpha$ in MG sampling. Setting $\alpha=40$ prevents objects from suddenly disappearing or emerging. However, excessively high $\alpha=150$ leads to over-smooth result.
    }
    \vspace{-1em}
    \label{fig:ablation}
\end{figure}


\begin{figure*}[!t]
    \centering
    \includegraphics[width=1\linewidth]{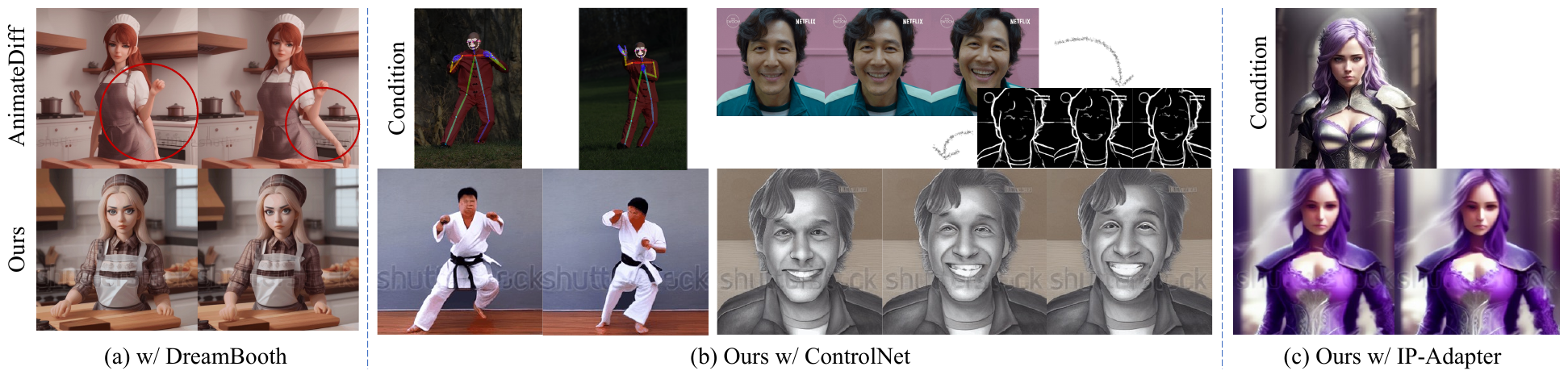}
    \caption{The use of a frozen StableDiffusion model enables the utilization of various off-the-shelf models without the need for retraining. (a) By employing Dreambooth/LoRa, Ours can generate specific domain videos. We yields more accurate motion and object representation compared to AnimateDiff. We use the Rcnzcartoon Dreambooth model. (b) ControlNet can be used to apply conditions to specific frames. (c) IP-Adapter allows for the generation of videos using images as conditions.
    }
    \vspace{-1em}
    \label{fig:dreambooth}
\end{figure*}

\fref{fig:ablation}a demonstrates effect of mapping network and frame-wise tokens. We extract two short vertical segments of fixed locations from 24 frames of each video, and stack them horizontally between the first and last frame. Comparing [A] and [B], the minimal variant of our method still bear inconsistent objects' appearance between frames, revealed by the first and last frames, while adding mapping network improves the consistency. Comparing [B] and [C], adding frame-wise tokens resolves flickering between adjacent frames which is demonstrated by jaggies and smoothness in the horizontal stack of the 24 frames, respectively. We recommend watching the video comparison in the supplementary materials.

\fref{fig:ablation}b visualizes the cross-attention maps of the tokens. While the text tokens affect similar region across different frames, the attention maps of the frame-wise tokens smoothly vary across frames. We suppose that the inflated networks soley cannot represent natural videos and the frame-wise tokens provide additional freedom to the model leading to more natural videos.

\fref{fig:ablation}c depicts the effect of varying the guidance scale $\alpha$ of mitigating gradient (MG) sampling.
MG sampling not only effectively reduces momentary artifacts but also enhances the overall quality of the video.
However, too large $\alpha$ may pause or distort the video.
We set $\alpha=40$ for the best results.

We provide a more detailed ablation study with additional videos in supplementary materials, demonstrating smoothness of motion and consistency of objects thanks to the proposed method. We encourage readers to observe qualitative aspects that are not captured by FVD.

\subsection{Personalized / image-conditioned videos}
Our model naturally supports combining existing off-the-shelf personalized T2I methods~\cite{ruiz2023dreambooth,hu2021lora,zhang2023adding,ye2023ip-adapter} because of the frozen pretrained T2I layers.
In \fref{fig:dreambooth}a, we compare personalized videos using DreamBooth~\cite{ruiz2023dreambooth} on AnimateDiff~\cite{guo2023animatediff} and ours. AnimateDiff produces spurious structure such as branching arm while our method produces videos with natural structure.
\fref{fig:dreambooth}b and \fref{fig:teaser} show the videos controlled by skeleton and sketches using ControlNet. We can either provide conditions of full frames or only a few non-adjacent frames, e.g., starting and ending frames. In the latter case, our model produces plausible inter-motion between the specified frames.
\fref{fig:dreambooth}c shows image-conditioned video using IP-Adapter. More results are in the supplementary materials.

This compatibility with off-the-shelf T2I variants is a huge advantage of our model leading to potential explosion of video generation era. Especially, our model does not require additional training to inflate the T2I variants of StableDiffusion.

\begin{table}[t]
\centering
\resizebox{0.6\linewidth}{!}{%
\begin{tabular}{lccccc}
\hline Method & Single stage & Frozen T2I & Training Param. &FVD $(\downarrow)$ & CLIPSIM ($\uparrow$) \\
\hline CogVideo (C) & \ding{55} & \ding{55} & - & 751.34 & 0.2614 \\
CogVideo (E) & \ding{55} & \ding{55} & - & 701.59 & 0.2631\\
Make-A-Video & \ding{55} & \ding{55} & 9.7B & 367.23 & 0.3049 \\
 VideoFusion$^\dagger$ & \yes{} & \ding{55} & - & 639.90 &  - \\
MagicVideo & \ding{55} & \ding{55} & - & 699.00 & - \\
 Video LDM & \ding{55} & \yes{} & 4.2B & 550.61 & 0.2929 \\
 LAVIE & \ding{55} & \ding{55} & 3B & 526.30 & 0.2949 \\
 PYOCO & \ding{55} & \ding{55} & 1.08B & 355.2 & - \\
 EMU VIDEO & \ding{55} & \ding{55} & 6B & 606.2 & - \\
 \hline StableDiffusion$^*$ & - & - & - & 2065.71 & - \\
  AnimateDiff$^{**}$ & \yes{} & \yes{} & 0.43B & 1934.72 & 0.2785 \\
 Ours & \yes{} & \yes{} & 0.45B & 787.87 & 0.2948 \\
\hline
\end{tabular}
}
\caption{\textbf{Quantitative comparison} using FVD on UCF101 and CLIP Similarity on MSR-VTT. AnimateDiff$^{**}$ is not for general Video, and StableDiffuion$^{*}$ result is from generating each frame independently. VideoFusion$^\dagger$ is trained on UCF101.}
\vspace{-1em}
\label{tab:comparison}
\end{table}

\begin{table}[!t]
\centering
\resizebox{0.8\linewidth}{!}{%
\begin{tabular}{c|cc|cc|cc}
Question & OURS & VideoLDM & OURS & PYoCo & OURS & ModelScopeV2 \\ \hline
Natural motion & \textbf{68.01\%} & 31.98\% & \textbf{74.07\%} & 25.92\% & \textbf{65.83\%} & 34.16\% \\
Consistency & \textbf{66.97\%} & 33.02\% & \textbf{76.47\%} & 23.52\% & \textbf{71.31\%} & 28.68\% \\
Quality & \textbf{66.15\%} & 33.84\% & \textbf{74.36\%} & 25.63\% & \textbf{71.61\%} & 28.38\% \\
\end{tabular}%
}
\caption{We conduct a user study to compare videos in aspects of motion, consistency, and quality. The participants rated ours higher than others in motion consistency, and overall quality.}
\vspace{-2em}
\label{tab:appendix_survey}
\end{table}

\subsection{Comparison to existing methods}
\tref{tab:comparison} showcases a comparative analysis with other studies in the field following the common practice. We emphasize that ours is the method with single-stage and frozen T2I layers for general video generation task. 
With a slightly different setting, ours is able to achieve FVD of 623.51, which is better than our standard setting, but noticed worse visual quality in general. Due to the instability of the FVD metric~\cite{emuvideo}, we stopped tuning our settings for a better FVD and aimed at the best visual quality. We prepared 100 unique text prompts with distinct movements (\tref{tab:eval_sentence}) and generated four videos per prompt. These 400 videos were then evaluated based on binary attributes to determine the best quality. Please refer to \tref{tab:additional-ablation}. A concurrent work, EmuVideo \cite{emuvideo}, also reports a relatively high FVD score and argues that automated metrics on UCF101 are flawed for zero-shot text-to-video evaluation. 

Additionally, the table presents FVD of StableDiffusion obtained by sampling each frame individually with class names from the UCF101 as text prompts. Our observations from the results of StableDiffusion highlight two key points as follows. First, the classes in UCF101 invariably include human hands and faces, which StableDiffusion struggles to accurately generate. Second, it tends to generate various domains such as illustrations and gray photographs. Our method, which employs frozen StableDiffusion, exhibits similar tendency. This, we believe, contributes to acheiving the relatively higher FVD. 

 Despite of FVD score, our method produces diverse and accurate movements while previous methods often fail, e.g., continuity of walking legs across frames. \tref{tab:appendix_survey} provides a user study with 50 participants with over 5K votes on 3 questions over 34 videos (13+13+8) randomly selected from the project pages of VideoLDM and PYoCo and generated by ModelScope. The participants rated ours higher than others in motion, consistency, and overall quality. We assert that our results are comparable to those of other large T2V models, despite our method being the only one with a single-stage process and frozen T2I layers.

\begin{figure*}[!t]
    \centering
    \includegraphics[width=0.8\linewidth]{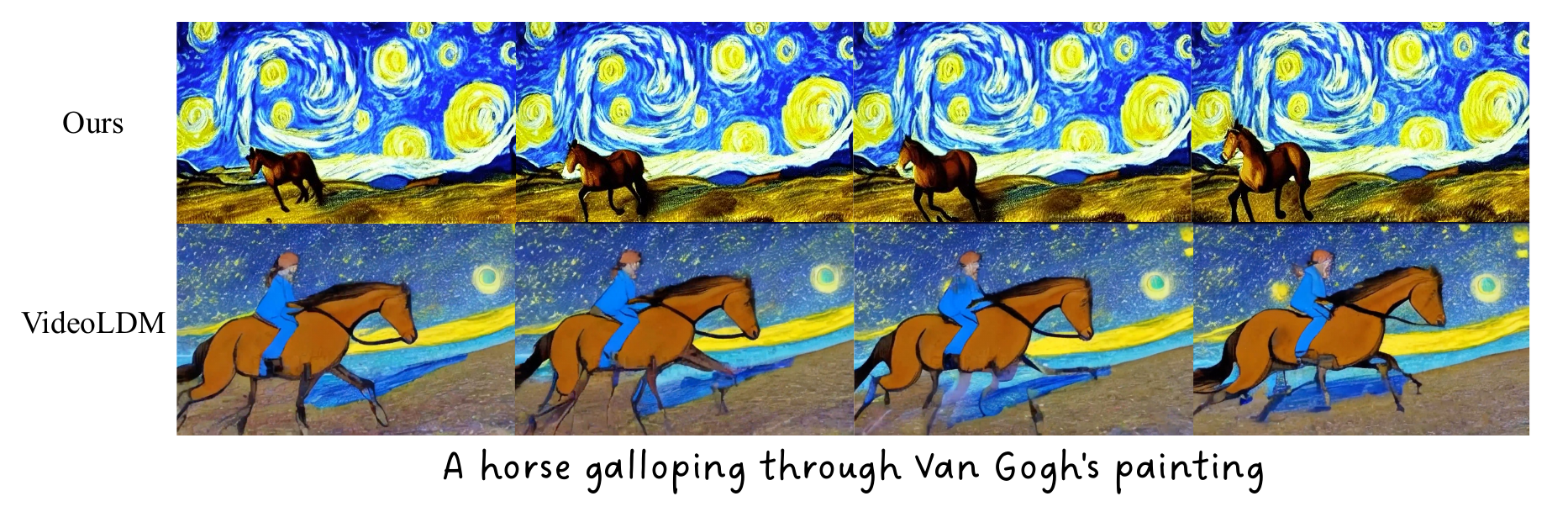}
    \centering
    \includegraphics[width=0.8\linewidth]{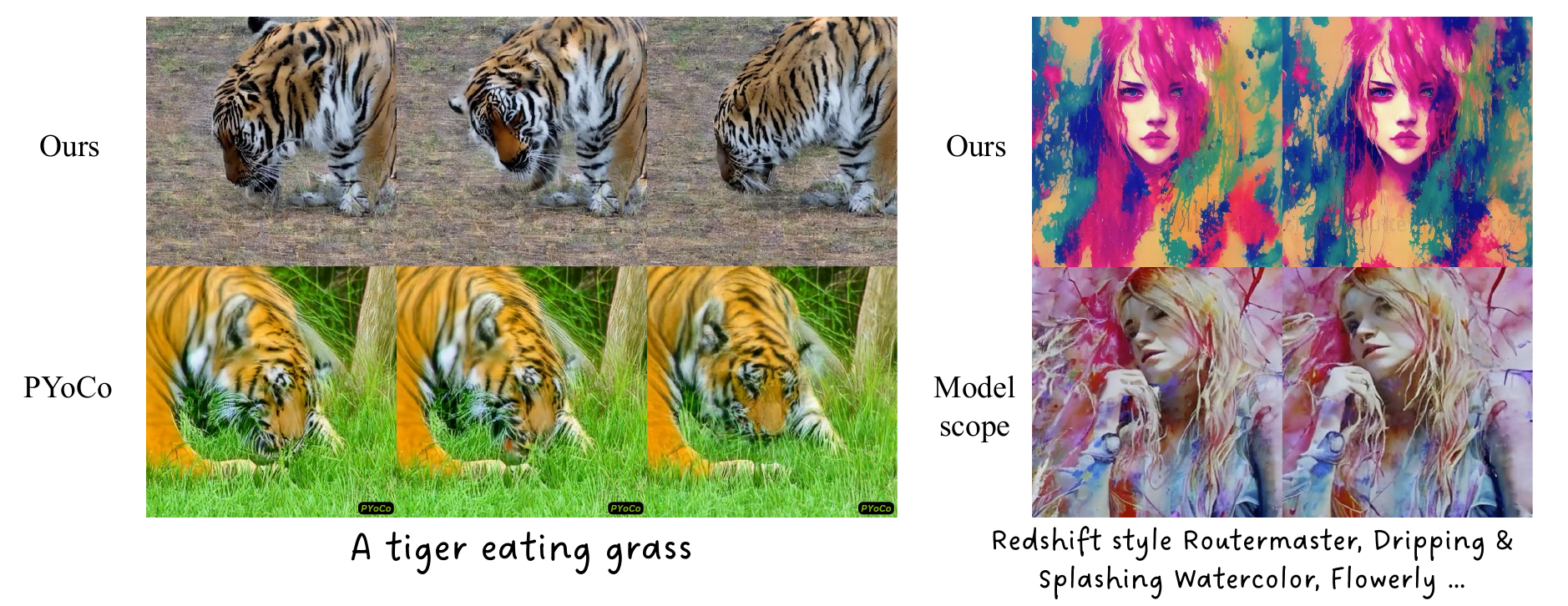}
    \caption{
    Our method generates diverse and accurate movements. Videos are from the project pages of VideoLDM and PYoCo and generated by ModelScope.
    }
    \vspace{-2em}
    \label{fig:compare12}
\end{figure*}

\section{Conclusion and Discussion}
This paper presents a video generation model built upon a pretrained text-to-image (T2I) model. It has novel network components including the mapping network and the frame-wise token generator and is supervised by novel losses such as temporal regularized self-attention loss, \vh{}-space contrastive loss, and image model regularization loss. Furthermore, it is accompanied by a mitigating gradient sampling that further improves smoothness and quality in generative process. All elements are motivated by inductive biases from characteristics of natural videos and contribute to the performance improvements. Last but not least, our model is efficient to train thanks to single-stage process and frozen T2I layers. We hope our method will provide an influential starting point for future T2V models and contribute to a healthy research community.

However, our T2V model has limitations due to its reliance on the StableDiffusion T2I model. This dependency means that issues in T2I models, like inaccurately generating human hands and limbs, also affect our T2V model. Exploring connections with other methods that have addressed these issues would be a valuable direction for future work.

It also raises ethical concerns related to video generation, such as the risks of misusing synthetic media. Our work underscores the importance of continuous improvement in generative models in video synthesis, while also highlighting the necessity for ethical considerations.

\clearpage




%
%
\bibliographystyle{splncs04}
\bibliography{main.bib}

\section*{Acknowledgments}
This work was supported by the National Research Foundation of Korea (NRF) funded by the Korean government (MSIT) (RS-2023-00223062).
\input{appendix}

\end{document}

%% file: macro.tex
\usepackage{amsmath,amsfonts,bm}

\newcommand{\fref}[1]{Figure~\ref{#1}}

\newcommand{\eref}[1]{Eq.~(\ref{#1})}
\newcommand{\tref}[1]{Table~\ref{#1}}

\def\temporal{$l_{\theta}$}
\def\mapping{$l^{M}_{\theta}$}
\def\spatial{$l_{\phi}$}

\def\trs{$L_{\text{TRS}}$}

\def\vx{\mathbf{x}}
\def\vh{$\mathbf{h}$}
\def\vz{\mathbf{z}}

\definecolor{darkgreen}{rgb}{0.0, 0.5, 0.0}
\def\yes{\textcolor{darkgreen}{\ding{51}}}

%% file: appendix.tex
\clearpage
\appendix

\setcounter{page}{1}

\renewcommand{\thetable}{A\arabic{table}}
\renewcommand{\thefigure}{A\arabic{figure}}
\setcounter{figure}{0}
\setcounter{table}{0}

\section{Network design}
\label{appendix:arch}
We provide pseudo-codes at the end of the appendix.

\paragraph{Temporal layers}
Similar to AnimateDiff \cite{guo2023animatediff}, we add temporal attention layers \temporal{} between spatial layers \spatial{} of the text-to-image model, as a common technique for T2V methods.
For each temporal layer \temporal{}, we reshape the input as $\texttt{((b f) c h w} \rightarrow \texttt{b c f h w)}$ and output as $\texttt{(b c f h w} \rightarrow \texttt{(b f) c h w)}$, where \texttt{b, c, f, h,} and \texttt{w} represent batch size, number of channels, number of frames, height, and width, respectively.

\paragraph{Mapping network}
Our mapping network $l_\theta^M$ tries to prepare the proper input for the text-to-video (T2V) U-Net. In order to consider the cross-frame relationship, we design its architecture to have 3D convolution layers for \texttt{f, h, w} dimensions and a temporal attention layer.
Initially, the input is reshaped from  $\texttt{(b f c h w} \rightarrow \texttt{(b h w) c f)}$, following which a 3D convolution is applied. Subsequently, the data undergoes processing through a temporal attention layer and a linear layer, facilitating effective noise distribution alignment within the frame sequence.

\paragraph{Token generator}
While typical architectures receive identical text tokens to cross-attention layers for all frames, our token generator $\boldsymbol{t_\theta}(\cdot)$ provides frame-wise text tokens that model detailed temporal variations beyond the standard text tokens of objects.
To facilitate this, the token generator transforms the original text tokens into frame-wise tokens. It utilizes a sequential network of linear and SiLU layers, followed by a 1D convolution layer. This structure is capable of expanding the text embeddings into a format suitable for representing each frame individually.

\paragraph{Projection layer $\boldsymbol{g}(\cdot)$}
We implement our projection layer $\boldsymbol{g}(\cdot)$ for the contrastive loss on $\vz$ space. It is an MLP with SiLU between linear layers of (\texttt{c h w})$\rightarrow$ \texttt{c}$\rightarrow$ \texttt{c} $\rightarrow$ \texttt{c} channels.


\section{Custom noise distribution of previous methods}
\label{appendix:mapping}
It is worth noting that the spatial layers of the Text-to-Image (T2I) model remain frozen, and we directly utilize them. This practice acknowledges the imbalance between the characteristics of a T2I model, trained to generate a variety of images starting from Gaussian noise, and the objectives of a Text-to-Video (T2V) model, which needs to create a sequence of images. PYoCo \cite{ge2023preserve} supposes that the IID noise prior is ill-suited for training video diffusion models due to the lack of temporal correlations between frames. Instead, they introduce a correlated noise model that uses specific shared noise across all frames to design frame correlations. However, this shared noise leads to a non-Gaussian distribution that contradicts the premise of the reparameterization trick, which is the direct conversion from $\vx_0$ to $\vx_t$. We contend that correlated noise across all frames cannot be considered Gaussian noise. As shown in Table \ref{tab:gaussian}, we generated shared noise 10,000 times to verify its Gaussian nature and found that only 63\% of correlated noises fit a Gaussian distribution according to the Jarque–Bera test~\cite{jarque1980efficient}. We also observe that such correlated noise leads to unstable training, particularly when using only video data.

\begin{table}[!t]
\centering
\begin{tabular}{c|cc}
Jarque–Bera         & Correlated noise \cite{ge2023preserve}   & Random noise   \\ \hline
p\textgreater{}0.05 & 63.01 \% & 94.42 \%
\end{tabular}
\caption{We randomly sampled 10,000 Correlated noise following PYoCo\cite{ge2023preserve} and random gaussian noise. According to the Jarque-Bera test with p\textgreater{}0.05, only 63\% of Correlated noises fit a Gaussian distribution which harms the assumption of the re-parameterization trick.}
\label{tab:gaussian}
\vspace{-1em}
\end{table}

\section{Effect of the mapping network}
\label{appendix:mapping}

\begin{figure}[!t]
    \centering
    \includegraphics[width=1.0\linewidth]{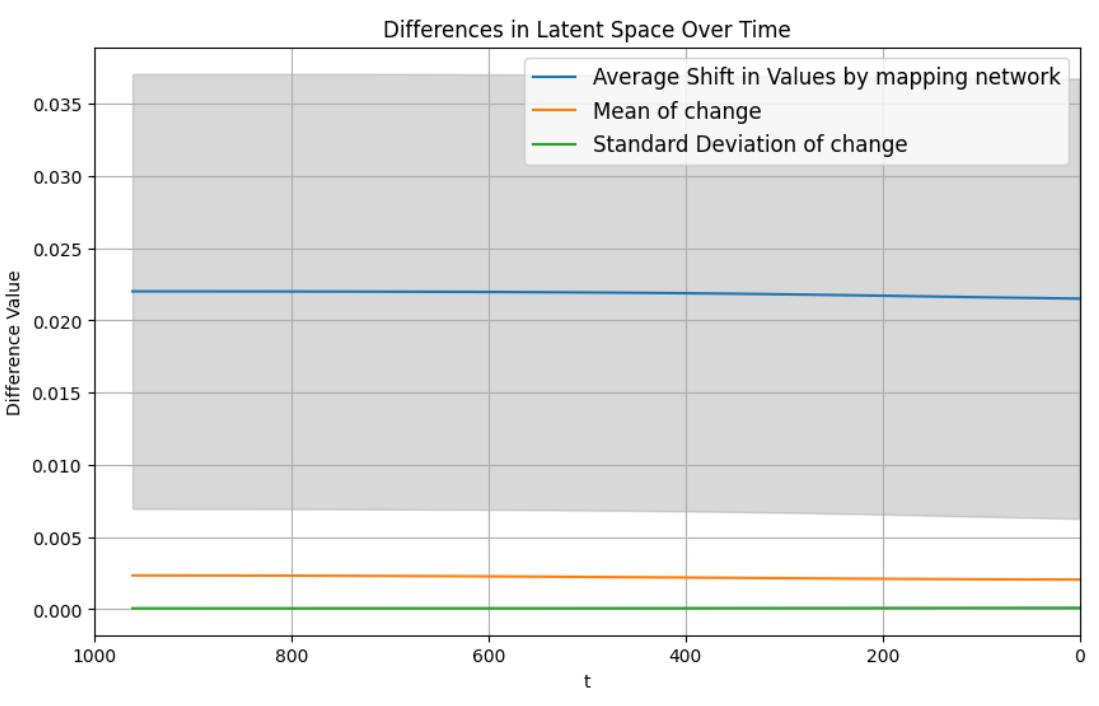}
    \vspace{-2em}
    \caption{The figure compares the data before and after passing through the mapping network across all sampling timesteps. The blue line represents the average of the absolute differences between the values before and after passing through the mapping network, and the gray area indicates a range of ±1 standard deviation. This indirectly demonstrates that significant changes have been applied after passing through the mapping network. The orange line shows the change in the mean value after passing through the mapping network, while the green line represents the change in standard deviation. Despite the significant additions made by the mapping network, the standard deviation remained largely unchanged, and a slight increase in the mean value was observed across all times.
    }
    \vspace{-1em}
    \label{fig:appendix_mapping}
\end{figure}

Our designed mapping network was implicitly engineered to modify an input distribution suitable for videos. \fref{fig:appendix_mapping} illustrates the small but non-negligible value changes observed after passing through the mapping network. The blue line depicts the mean of the absolute differences between values before and after the mapping network, while the gray area signifies a range of ±1 standard deviation. This suggests that considerable modifications have been implemented post-mapping network passage. The orange line indicates the mean value shift after mapping, and the green line shows the standard deviation change. Notably, even with alterations by the mapping network, the standard deviation mostly remained stable, with only a minor increase in mean values across all times.

\begin{figure}[!t]
    \centering
    \includegraphics[width=0.7\linewidth]{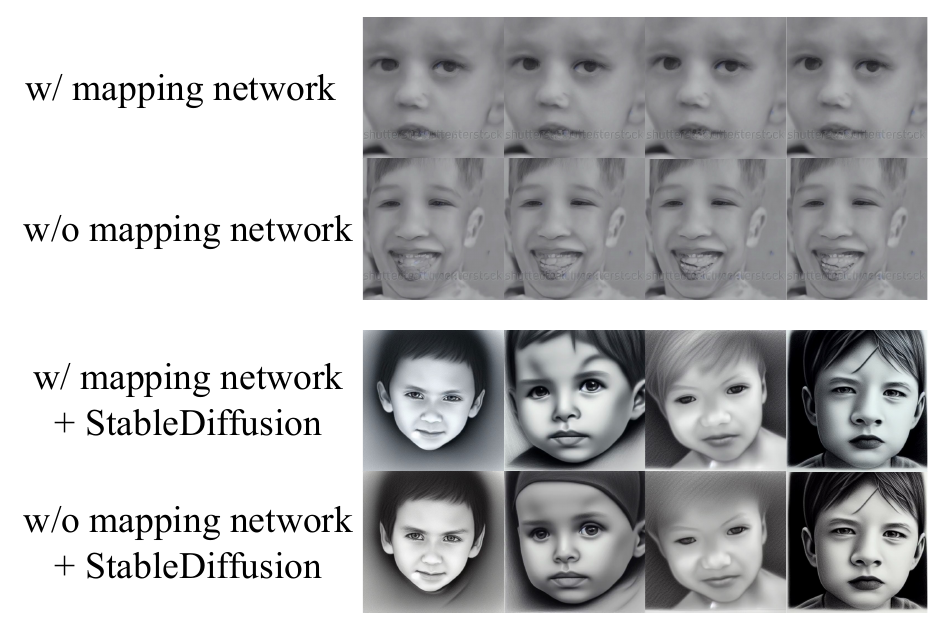}
    \vspace{-1em}
    \caption{When generating videos without the mapping network in our final model, we observed that the content changes completely, and the videos are distorted. However, intriguingly, when used in conjunction with StableDiffusion, we observed that similar images of the same content are produced regardless of whether the mapping network is included or not.
    }
    \vspace{-1em}
    \label{fig:appendix_mapping2}
\end{figure}

\fref{fig:appendix_mapping2} displays the outcomes when the mapping network is not utilized. Since the model was trained with the mapping network, omitting it artificially results in changed and corrupted content. Interestingly, when StableDiffusion is employed with the mapping network, the end results are nearly identical to those without it. We speculate this demonstrates that the mapping network implicitly modifies a distribution without notably impacting the image model.

\section{More details}
\label{appendix:details}
Here we share more detailed settings.

\paragraph{Training} 
We initiated our training with a learning rate of 0.0001 and gradually decreased it to 0.00001. We adopted a beta start of 0.00085 and an end of 0.012, utilizing a linear beta schedule. The training process took approximately one or two weeks on 32 A100 GPUs. We separately trained models for 256$^2$ resolutions, with all numerical data and experiments in the paper conducted using the 256$^2$ model. However, we also found that models trained on 256$^2$ resolution videos could generate up to 512$^2$ resolutions. Notably, in cases where the text prompts were very short or abstract, the 256 model struggled to generate satisfactory 512$^2$ resolution outputs. (e.g., with text prompt "A man", generated video usually contains more than two men.) We observed that training with 512$^2$ resolution videos also trained well but required more time.

\paragraph{Inference} We used the DDIM scheduler for inference, sampling all videos with 25 steps. We observed that using a higher classifier free guidance (CFG) scale was effective for generating imaginative and unrealistic videos. Particularly, we applied stronger guidance as t approached T. We used a CFG scale of 10-15 up to 0.7T and 7.5 thereafter. For MG sampling, we adopted a scale of 40-50, noting that some of the attached videos were generated using $\alpha$=50.

MG sampling is executed in closed form as per Eq. 6 and the cost is negligible. Also it impacts greatly when near noise by $\omega$, as in the attached rabbit video, helping to enhance and increase appropriate motion. Also, it is not sensitive at all if not using $\alpha>$200.

\section{More ablation study}
\label{appendix:ablation}

To measure our model's performance, we created 100 custom prompts and evaluated them across four categories (See \tref{tab:eval_sentence}.) The prompts were evenly distributed into five themes: humans, animals, objects, nature, and complex/imaginary, each containing 20 sentences. All sentences were composed of clear, simple language with verbs indicating distinct actions or movements. For each prompt, we sampled four videos to check if the content matched the text, if there was action aligned with the verb, how smooth the video was, and whether there were any strange artifacts or collapsing objects.

We chose this approach for several reasons: Firstly, measuring FVD requires sampling over 10,000 videos, which is time-consuming. Secondly, FVD does not adequately reflect the movement or motion in videos. Thirdly, the classes in UCF101 are too limited to effectively evaluate a text-to-video model. Therefore, we based all our experiments on the 100 prompts we designed and conclude that all the components we proposed significantly enhance performance as evident in \tref{tab:additional-ablation}. 

\paragraph{Current Limitations of Qualitative Metrics in Video Generation}
The current qualitative metrics do not adequately reflect video quality. EmuVideo\footnote{\scriptsize https://arxiv.org/pdf/2311.10709} and STREAM\footnote{\scriptsize (ICLR2024) https://arxiv.org/pdf/2403.09669} argue that these metrics are not aligned with human evaluations and are merely extensions of those for image generative models. Additionally, UCF101’s limited classes are insufficient for effective evaluation of video models, generating videos from simple words like “Hula Hoop,” which is unsuitable for text-to-video generation. Recognizing these limitations, the community is developing new metrics like STREAM. We believe our 120+ diverse videos demonstrate video quality beyond quantitative results.

\begin{table}[t]
\centering
\scriptsize
\begin{tabular}{l|cccc|c}
   \%       & content & action & smooth & non-weird  & AVG   \\ \hline
Basline   & 41                   & 30                  & 16           & 14                     & 25.25 \\
+ $L_{\text{TRS}}$   & 72                   & 60                  & 60           & 54                    & 61.5  \\
+ $L_{\text{DC}}$     & 76                   & 61                  & 66           & 69                    & 68    \\
+ Mapping network   & 81                   & 63                  & 73           & 69                  & 71.5  \\
+ Framewise tokens & 81                   & 68                  & 81           & 76                     & 76.5  \\
+ MG sampling       & 82                   & 68                  & 87           & 85                   & 80.5
\end{tabular}
\caption{
We generated videos using 100 unique text prompts and then conducted a binary evaluation for each prompt. The content category represents text-aligned content, while the action category indicates verb-aligned action. For the non-weird category, we counted videos without artifacts or distortions in objects. We noticed that the WebVid-10M captions largely lack camera motion and movement direction, hence we did not check for camera/object movement direction alignment.}
\vspace{-1em}
\label{tab:additional-ablation}
\end{table}

\section{More samples}
\label{appendix:samples}

\textbf{Please see the project page.}

\clearpage

\begin{longtable}{c|l}
\caption{100 evaluation sentences} \label{tab:eval_sentence} \\
category & sentences \\ \hline
\endfirsthead
\multicolumn{2}{c}%
{{\tablename\ \thetable{} -- continued from previous page}} \\
category & sentences \\ \hline
\endhead
\hline \multicolumn{2}{r}{{Continued on next page}} \\ \hline
\endfoot
\hline
\endlastfoot
human     &  A black man is walking left to right.  \\ 
          & A white man is yawning.  \\ 
          & An Asian man is painting on a canvas.  \\ 
          & A man drives a car.  \\ 
          & A man is playing the guitar, camera zoom-in.  \\ 
          & A black woman is running left to right.  \\ 
          & A white woman is falling down.  \\ 
          & An Asian reads a book, camera zoom-in.  \\ 
          & A woman is playing tennis.  \\ 
          & A woman is playing boxing.  \\ 
          & A baby is crawling, getting closer.  \\ 
          & A baby is sucking a bottle.  \\ 
          & A boy swings an arm.  \\ 
          & A boy drinks the orange juice.  \\ 
          & A girl is jumping right to left.  \\ 
          & A girl is eating a slice of pizza.  \\ 
          & A couple is dancing.  \\ 
          & A couple is toasting wine, camera zoom-out.  \\ 
          & Children are playing.  \\ 
          & A group of children is writing on chalkboard. \\ \hline
animals   & A dog is running left to right. \\ 
          & A big black dog barks. \\ 
          & A cat is jumping right to left. \\ 
          & A white cat chases a ball, moving away. \\ 
          & A flamingo is wading right to left. \\ 
          & The flamingo eats fish. \\ 
          & A zebra is galloping, moving away. \\ 
          & A zebra eats grass, camera zoom-in. \\ 
          & An eagle is soaring, getting closer. \\ 
          & An eagle is sitting on a tree, camera zoom-in. \\ 
          & A shark is swimming left to right. \\ 
          & A shark is struggling in a net. \\ 
          & A spider is crawling right to left. \\ 
          & A spider is hanging from a drizzled web. \\ 
          & A snake is slithering, getting closer. \\ 
          & A yellow snake is hunting a rat. \\ 
          & A clownfish is swimming. \\ 
          & A clownfish is hiding in the sea anemone. \\ 
          & A butterfly is fluttering. \\ 
          & A butterfly is sucking the flower. \\ \hline
objects   & The soccer ball is bouncing, moving away. \\ 
          & The windmill is spinning, camera zoom-out. \\ 
          & The helicopter is hovering, camera zoom-out. \\ 
          & The rocket is launching. \\ 
          & The pendulum is swinging. \\ 
          & The airplane is flying right to left. \\ 
          & The ship is sailing right to left. \\ 
          & The car is speeding. \\ 
          & The train is going left to right. \\ 
          & The kite is flying, camera zoom-out. \\ 
          & The clock is ticking, camera zoom-in. \\ 
          & The fireworks are exploding. \\ 
          & The light in the room is flickering. \\ 
          & The glass is breaking. \\ 
          & The alarm is ringing. \\ 
          & The popcorn is popping. \\ 
          & An apple is falling down from the tree. \\ 
          & A satellite is floating in space. \\ 
          & A school bus is stopping, camera zoom-out. \\ 
          & A paper airplane is flying in the sky. \\ \hline
nature    & Clouds are floating left to right. \\ 
          & Waterfalls are flowing, camera zoom-in. \\ 
          & Trees are swaying. \\ 
          & Flower fields are blooming. \\ 
          & Waves are splashing on beaches. \\ 
          & The sun is shining. \\ 
          & Fallen leaves are dropping. \\ 
          & Lightning is flashing. \\ 
          & Rain is falling. \\ 
          & Snow is falling. \\ 
          & The river is flowing. \\ 
          & The stars are twinkling. \\ 
          & The mist is rising. \\ 
          & A tree is burning. \\ 
          & Volcanoes are erupting, camera zoom-out. \\ 
          & Bamboo is rustling. \\ 
          & Reeds are swaying. \\ 
          & Gales are sweeping. \\ 
          & Seeds are germinating. \\ 
          & A glacier collapses. \\ \hline
complex \\ imaginary & A panda is eating pizza in front of the waterfalls. \\ 
          & A monkey is riding a bicycle in the green park. \\ 
          & A tiger in a space suit is walking on the moon. \\ 
          & Giant ants dance with headsets on. \\ 
          & A red car is flying to the space. \\ 
          & A dog and cat are playing with rolling ball. \\ 
          & A man in a suit is staring into the distance with a black umbrella \\ 
          & \quad under a streetlamp on a rainy night. \\ 
          & A lion in a suit opens the door and comes in. \\ 
          & A giant meteor is falling from the sky. \\ 
          & A giant house-sized doughnut is rolling on the road. \\ 
          & A dragon is shooting fire in the sky. \\ 
          & The man in the European painting on the wall greets with a wave. \\ 
          & The dove keeps coming out of the magician's hat. \\ 
          & The woman in the dress is slowly descending \\ 
          & \quad from the sky in a black umbrella. \\ 
          & A giant mantis is threatening people in the forest. \\ 
          & A flying train is flying in the night sky against \\ 
          & \quad the backdrop of the full moon. \\ 
          & Pegasus is drinking from the river. \\ 
          & Centaur is running fast through the field with a spear. \\ 
          & Aurora is dancing in the night sky with two moons. \\ 
          & The cat is transforming into a human. \\ 
\end{longtable}

\section*{Mapping layer}
\begin{lstlisting}[language=python]
class MappingLayer(nn.Module):
    def __init__(self, query_dim: int, heads: int = 8, dim_head: int = 64):
        super().__init__()
        self.temporal_attn = CrossAttention(query_dim=query_dim, heads=heads, dim_head=dim_head)
        self.temporal_norm = nn.LayerNorm(query_dim)
        self.proj_out = zero_module(nn.Linear(query_dim, query_dim))
        self.temporal_conv3d_residual = TemporalConv3d_Residual(query_dim)
        
    def forward(self, x, pass_motion_module=False):
        if pass_motion_module:
            pass
        b, c, f, hs, ws = x.shape
        input_x = x
        x = rearrange(x, "b c f h w -> (b h w) c f", b=b, f=f).contiguous()
        x = self.temporal_conv3d_residual(x)
        x = rearrange(x, "bhw c f -> bhw f c").contiguous()
        h = self.temporal_attn(self.temporal_norm(x))
        h = self.proj_out(h)
        h = h + x
        h = rearrange(h, "(b h w) f c -> b c f h w", b=b, h=hs, w=ws).contiguous()
        return h + input_x
\end{lstlisting}

\section*{Contrastive projection layer}
\begin{lstlisting}[language=python]
class Contrastive_projection(nn.Module):
    def __init__(self, ch, ds, input_video_resolution):
        super(Contrastive_projection, self).__init__()
        kernel_size = input_video_resolution // ds
        self.layers = nn.Sequential(
            nn.Conv2d(ch, ch, kernel_size=kernel_size, stride=1, padding=0, bias=False),
            nn.SiLU(),
            nn.Flatten(),
            linear(ch, ch),
            nn.SiLU(),
            linear(ch, ch),
            nn.SiLU(),
            linear(ch, ch)
        )
        
    def forward(self, x):
        return self.layers(x)
\end{lstlisting}

\section*{Frame-wise token generator}
\begin{lstlisting}[language=python]
class Frame_wise_token(nn.Module):
    def __init__(self, text_embed_dim, max_text_len, framewise_cond_token_num, num_frames):
        super(Frame_wise_token, self).__init__()
        self.f = num_frames
        self.frame_wise_token_num = framewise_cond_token_num
        if self.frame_wise_token_num > 0:
            self.layers = nn.Sequential(
                self._linear_silu(text_embed_dim, text_embed_dim),
                nn.SiLU(),
                nn.Conv1d(in_channels=max_text_len, out_channels=framewise_cond_token_num * num_frames, kernel_size=1, stride=1, padding=0, bias=False),
                nn.SiLU(),
                self._linear_silu(text_embed_dim, text_embed_dim),
                self._linear_silu(text_embed_dim, text_embed_dim),
                self._linear_silu(text_embed_dim, text_embed_dim),
                self._linear_silu(text_embed_dim, text_embed_dim)
            )
            
    def _linear_silu(self, in_features, out_features):
        return nn.Sequential(linear(in_features, out_features), nn.SiLU())
        
    def forward(self, x, pass_motion_module=False):
        if self.frame_wise_token_num > 0 and not pass_motion_module:
            frame_wise_token = self.layers(x)
            frame_wise_token = rearrange(frame_wise_token, "b (f n) c -> b f n c", f=self.f)
            x = repeat(x, "b n c -> b f n c", f=self.f)
            x = torch.cat([x, frame_wise_token], dim=2)
            x = rearrange(x, "b f n c -> (b f) n c")
        else:
            x = repeat(x, "b n c -> (b f) n c", f=self.f)
        return x
\end{lstlisting}